\pdfoutput=1

\documentclass[11pt]{article}

\usepackage[]{latex/acl}

\usepackage{times}
\usepackage{latexsym}

\usepackage[T1]{fontenc}

\usepackage[utf8]{inputenc}

\usepackage{microtype}

\usepackage{inconsolata}

\usepackage{inconsolata}
\usepackage{latexsym}
\usepackage{float}
\usepackage{mathrsfs}
\usepackage{amsthm,amsmath,amssymb}
\usepackage{multirow}
\usepackage[ruled]{algorithm2e}
\usepackage{threeparttable}
\usepackage{array}
\usepackage{booktabs}
\usepackage{bm}
\usepackage{graphicx}
\usepackage{amssymb}
\usepackage{wasysym}
\usepackage{bbm}

\definecolor{Mycolor1}{HTML}{FDC3D2}
\definecolor{Mycolor2}{HTML}{DCFDC4}
\title{Forgetting before Learning: Utilizing Parametric Arithmetic for Knowledge Updating in Large Language Models}

\author{
Shiwen Ni{$^{\clubsuit}\footnotemark[1]$}, Dingwei Chen{$^{\spadesuit}\thanks{~~Equal contribution.}$}, Chengming Li{$^{\diamondsuit\dagger}$}, Xiping Hu{$^{\diamondsuit}$}, {\bf Ruifeng Xu}$^{\heartsuit}$, \textbf{Min Yang}$^{\clubsuit}\thanks{~~Corresponding author.}$\\
$^\spadesuit$ Sun Yat-Sen University  \\
$^\clubsuit$ Shenzhen Institutes of Advanced Technology, Chinese Academy of Sciences \\
$^\diamondsuit$Shenzhen MSU-BIT University 
~~~~$^\heartsuit$ Harbin Institute of Technology (Shenzhen)\\
  \texttt{\{sw.ni, min.yang\}@siat.ac.cn}, chendw26@mail2.sysu.edu.cn\\
  \texttt{\{licm, huxp\}@smbu.edu.cn, xuruifeng@hit.edu.cn}
  }
  
\begin{document}
\maketitle
\begin{abstract}
Recent advancements in Large Language Models (LLMs) have showcased their remarkable capabilities in text understanding and generation. However, even stronger LLMs are susceptible to acquiring erroneous or obsolete information from the training corpus. Direct secondary fine-tuning with data containing new knowledge may be ineffective in updating knowledge due to the conflict between old and new knowledge. In this paper, we propose a new paradigm for fine-tuning called \textbf{F-Learning} (\underline{F}orgetting before \underline{Learning}), which employs parametric arithmetic to facilitate the forgetting of old knowledge and learning of new knowledge. Experimental results on two publicly available datasets demonstrate that our proposed F-Learning can obviously improve the knowledge updating performance of both full fine-tuning and LoRA fine-tuning, simultaneously outperforming the existing baselines in most cases. Moreover, we have also discovered that forgetting old knowledge by subtracting the parameters of LoRA can yield a similar effect to subtracting the parameters of full fine-tuning, and occasionally even surpass it significantly.
\end{abstract}

\section{Introduction}
Large Language Models (LLMs) possess an extraordinary ability to understand and generate natural language \cite{brown2020language, raffel2020exploring, ouyang2022training}. Although LLMs are very capable of learning, they are not immune to the acquisition of incorrect knowledge in the corpus. Moreover, much of the knowledge in the real world is constantly updated, and some of the originally correct knowledge in LLMs can become outdated and invalid over time. For example, the question "Who is the President of the United States? is answered "Donald Trump" in the year 2020, while the answer now is "Joe Biden". Consequently, the challenge with LLMs is continuously updating to ensure they reflect current, correct knowledge. Existing methods of model editing and knowledge updating usually add additional network \citep{dong2022calibrating, huang2022transformer, raunak2022rank}, model parameters \citep{dai2023neural,dong2022calibrating,huang2022transformer}, knowledge bases \citep{murty2022fixing, mitchell2022memory, li2022large, madaan2022memory,mitchell2022memory,zheng2023can}, etc., and the editing process is not as straightforward and simple as fine-tuning methods \citep{zhang2022adaptive,li2021prefix,hu2021lora} directly with new knowledge. Currently, the most used method for learning new knowledge is still direct fine-tuning of the model.
\begin{figure}[t]
	\centering
\includegraphics[width=1\linewidth]{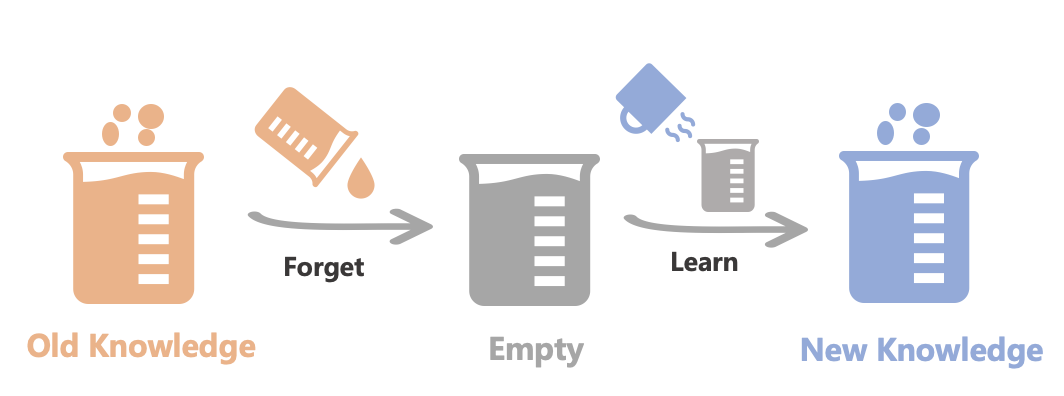}
	\caption{Diagram for “Forgetting before Learning”.
	}
	\label{f1}
\end{figure}

Empirically, when human beings establish their own initial cognition, if they are exposed to new knowledge that is inconsistent with their initial cognition, they usually feel conflicted and it is difficult for them to learn and accept the new knowledge. If the original cognition and knowledge are forgotten, then the new knowledge to be learned will not conflict with the original cognition and knowledge, which makes it better to learn and absorb the new knowledge. As shown in Figure \ref{f1}, it is better to pour in the "new water" only after the "original water" in the cup has been poured out. For example, if people have been educated to believe that "the Earth is flat" since childhood, it would be challenging for them to accept the conflicting knowledge that "the Earth is round" when they become adults. Conversely, if they can forget the erroneous knowledge that "the Earth is flat" or if they learn the correct knowledge that "the Earth is round" before being exposed to the incorrect information, it would be much simpler.

Inspired by the above empirical observations and \citep{ilharco2022editing}'s task arithmetic, we propose a novel paradigm of knowledge updating called \textbf{F-Learning} (\underline{F}orgetting before \underline{Learning}). Specifically, we first fine-tune the initial model using old knowledge and then subtract the difference between the fine-tuned model parameters and the initial model parameters from the initial model parameters. This process is defined as "\textit{old knowledge forgetting}". We then use the new knowledge to fine-tune the model after forgetting the old knowledge. This process we define as "\textit{new knowledge learning}". After the two stages of \textit{forgetting old knowledge} and \textit{learning new knowledge}, the model's knowledge is updated. The contribution of this work can be summarised as follows:

\begin{itemize}
    \item We propose a novel fine-tuning paradigm “Forgetting before Learning” (F-Learning) for knowledge updating in large language models. 
    \item Experimental results show that our proposed F-Learning improves the knowledge updating performance of various fine-tuning methods and outperforms the existing baselines in most cases.
    \item Experimental results show that forgetting by subtracting the parameters of LoRA can achieve the approximate effect of subtracting the parameters of full fine-tuning.
\end{itemize}

\section{Related Work}

Currently, the method of knowledge updating and model editing (also known as knowledge editing) for LLMs is mainly divided into two classes \cite{yao2023editing, wang2023knowledge}: 

\paragraph{\textit{a. The method preserving model's parameters}}

\textbf{Retrieve augmentation} practically depends on an external knowledge base which contains new or correct knowledge. Aiming at amending the output of LLMs, a new knowledge base will be connected with the base model to implement a retrieve for needed new knowledge to a prompt or a question \cite{murty2022fixing, mitchell2022memory, li2022large, madaan2022memory}. 
Mitchell et al. \cite{mitchell2022memory} store manual edits in a memory module, and use a classifier to call the knowledge stored in the memory. Madaan et al. \cite{madaan2022memory} leverage the memory of user's feedback to generate prompts for LLMs. Instead of gradient calculation, Zheng et al. \cite{zheng2023can} utilize the in-context learning method to revise the output of LLMs with demonstrations extracted from the corpus based on similarity. 

\noindent \textbf{Adding Additional Parameters} refers to injecting a few trainable parameters which represent new knowledge to LLMs while original parameters keeping frozen \cite{dong2022calibrating, huang2022transformer, raunak2022rank, dai2023neural}. Dong et al. \cite{dong2022calibrating} put forward a lightweight feed-forward network to add new parameters adapted to specific factual contexts for knowledge generalization. Huang \cite{huang2022transformer} et al. design an editor called Transformer-Patcher, which is capable of modifying the mistake of LLMs sequentially by adding and training a few neurons in transformer.

\paragraph{\textit{b. The method modifying model's parameters}}
\textbf{Fine-tuning} is a general technique since pre-training model has been widely adopted in NLP research, which always obtains promising results in downstream tasks. Meanwhile, fine-tuning is an intuitive and effective method to urge the model to learn new knowledge for model editing \cite{zhu2020modifying, zhang2022adaptive, yao2023editing}. Recently, there are a series of parameter-efficient fine-tuning methods, such as Prefix-Tuning \cite{li2021prefix} and LoRA \cite{hu2021lora}, making it more appreciate for knowledge editing based on fine-tuning. Zhang et al. \cite{zhang2022adaptive} operate incremental parameter updates of different amounts by calculating the importance of the weight matrix to improve the update efficiency and adaptability. Zhu et al. \cite{zhu2020modifying} leverage a loss constraint attached to the base model to reduce the impact on irrelevant knowledge during the process of fine-tuning. Similarly, Lee et al. \cite{lee2022plug} also implement large-scale continual learning for knowledge updating with regularized fine-tuning.

\noindent \textbf{Meta-learning} is aimed at updating the knowledge in LLMs through varying their parameters with the prediction from a well-trained hypyernetwork \cite{sinitsin2019editable, mitchell2021fast, de2021editing}. Mitchell et al. \cite{mitchell2021fast} propose an auxiliary network with gradient decomposition, which can execute efficient edits to LLMs according to a single input-output pair. De Cao et al. \cite{de2021editing} update part of weights for a subset of modules in the model relying on a hypernetwork with constrained optimization.

\noindent \textbf{Locate and edit} is related to the internal mechanism of the LLMs. With the help of some attributes, it usually locates the parameters and neurons in the light of specific knowledge and modifies them to correct the output \cite{meng2022locating, dai2022knowledge, meng2022mass, santurkar2021editing, geva2022transformer}. Geva et al. \cite{geva2021transformer} find that the feed-forward networks layer of the transformer stores key-value pairs which are related to specific knowledge. Meng et al. \cite{meng2022locating} utilize a causal reasoning method to distinguish the key neuron activations and update specific factual associations by modifying feed-forward weights. Furthermore, to implement knowledge editing on a large scale, they put forward MEMIT \cite{meng2022mass}, a method that directly updates thousands of memories in LLMs. Gupta et al. \cite{gupta2023editing} improve the knowledge updating through varying edit tokens and ameliorating the layer selection during the editing process. Yu et al. \cite{yu2023unlearning} leverage the partitioned gradient to identify the significant weights for unlearning of bias in the model.

In conclusion, there are many ways to achieve knowledge updating, but most of them require the addition of additional knowledge bases, neural network modules, and model parameters, which are cumbersome in practice and increase inference consumption. \textbf{This paper focuses on the improvement and enhancement of fine-tuning methods.}

\begin{figure*}[t]
	\centering
\includegraphics[width=1\linewidth]{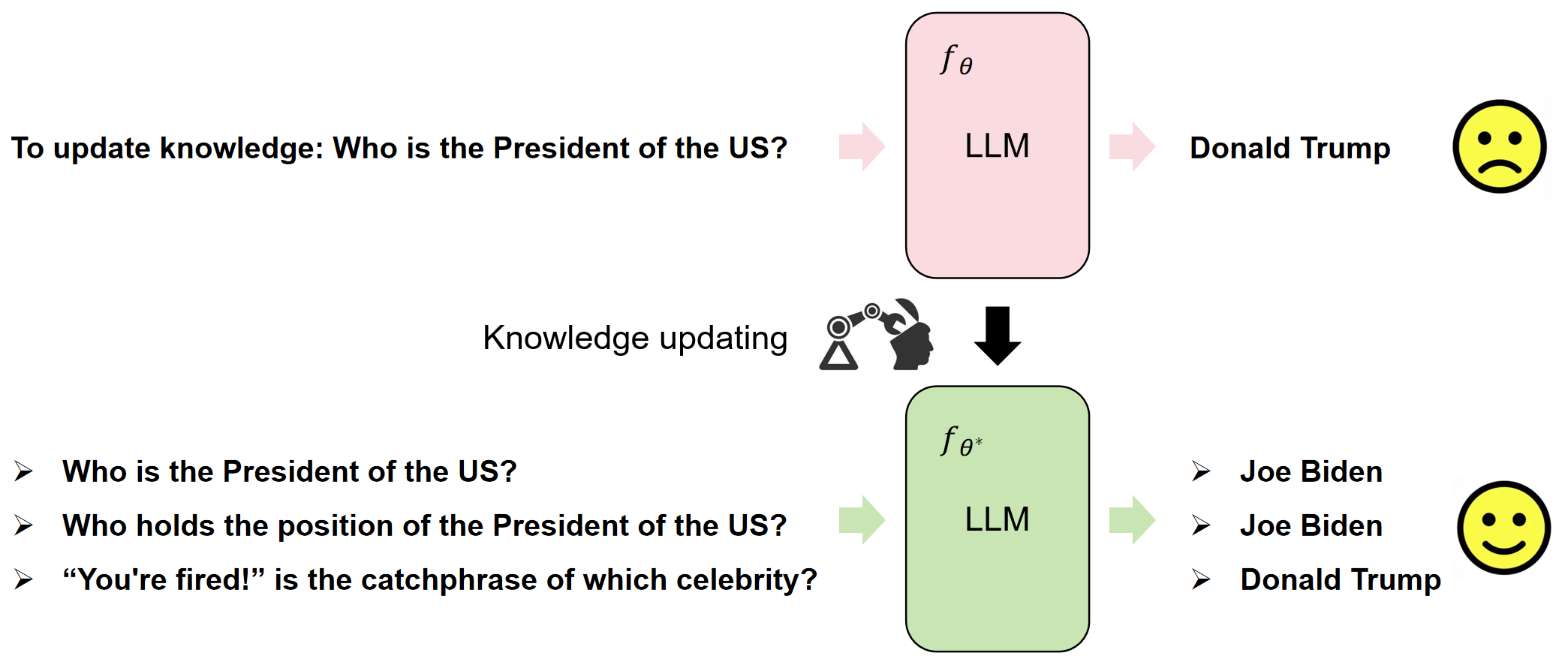}
	\caption{Objectives of the knowledge updating in large language model.
	}
	\label{f2}
\end{figure*}
\section{Task Definition}
Our task is knowledge updating of large models, which can be defined as given a model $f_{\theta}$ and a set of input-output knowledge pairs $K_{old}=\{(x_1,y_1),(x_2,y_2), ... ,(x_i,y_i)\}$, the parameters of the model need to be edited to obtain a new model $f_{\theta^{*}}$ (x) and a corresponding set of new input-output pairs $K_{new}=\{(x_1,y^{new}_1),(x_2,y^{new}_2),... ,(x_i,y^{new}_i)\}$. The $i$ is the number of knowledge pairs to be updated.
Referring to \citep{yao2023editing}, we can define this process and objective of knowledge updating as:
\begin{equation}\label{key}
	f_{\theta^{*}}(x_i)= \begin{cases}
y^{new}_i & \text{if } x_i \in N(x_i) \\
f_{\theta}(x_i) & \text{if } x_i \in other
\end{cases}
\end{equation}
where $N(x_i)$ represents $x_i$ itself and its equivalent neighbourhood. The knowledge update task needs to update only the answers of $x_i$ itself and its equivalent domain $N(x_i)$ without changing the answers of other out-of-scope knowledge. Specifically, the quality of knowledge updating has the following three evaluation indicators: (1) \textbf{Reliability} is measured as the average accuracy on the new knowledge for the updated model $f_{\theta^{*}}$. It's the first indicator of the effectiveness of knowledge updating. As shown in Figure \ref{f2}, the output of the question “Who is the President of the US?” needs to be updated from “Donald Trump” to “Joe Biden”. (2) \textbf{Generalization} means the new model $f_{\theta^{*}}$ should also updated the equivalent neighbour $N(x_i)$ (e.g. rephrased sentences). It is evaluated by the average accuracy of the model $f_{\theta^{*}}$ on examples drawn uniformly from the equivalence neighborhood. As shown in Figure \ref{f2}, the output of the question "Who holds the position of the President of the US?" also needs to be updated from "Donald Trump" to "Joe Biden". (3) \textbf{Locality} means the updated model $f_{\theta^{*}}$ should not change the output of the irrelevant examples. Hence, the locality is evaluated by the rate at which the updated model $f_{\theta^{*}}$'s predictions are unchanged as the pre-update $f_\theta$ model. As shown in Figure \ref{f2}, the output of the question “'You're fired!' is the catchphrase of which celebrity?” is to be kept unchanged as “Donald Trump”.

\section{Proposed method: F-Learning} \label{Proposed method: F-Learning}

In this section, we will present our method of knowledge updating for LLMs. Instead of introducing an external knowledge base or additional parameters, our method is mainly based on \textbf{full fine-tuning} and \textbf{parameter-efficient fine-tuning}. Briefly, it consists of two stages: 

\subsection{Forgetting old knowledge}

The supervised fine-tuning (SFT) on a dataset injects new knowledge into the LLMs or activates their fitting capabilities related to the new knowledge, which is reflected in the variation of the model's parameters. During this stage, for a given large language model $f_{\theta}$ and its parameters $\theta$, we define the incremental parameters as \textbf{knowledge parameters} $\theta_\vartriangle$, calculated as follows:

\begin{equation}\label{key}
    \theta_\vartriangle=\rm{FT}\{\theta,K \} - \theta
\end{equation}

\noindent where $\rm{FT}$ is the operation of supervised fine-tuning, while $K$, $\theta$ refer to the dataset of knowledge and the parameters of the original model $f_{\theta}$, respectively. Similarly, we first fine-tune the model $f_{\theta}$ on a dataset containing old knowledge, and then subtract the parameters $\theta$ of the original model $f_{\theta}$ from model's parameters after fine-tuning to obtain the knowledge parameters $\theta^{old}_\vartriangle$ indicating the old knowledge, as follows:

\begin{equation}\label{key}
    \theta^{old}_\vartriangle=\rm{FT}\{\theta,K_{old} \}- \theta,
\end{equation}
\noindent where $K_{old}$ refers to a dataset composed of old knowledge which we desire to forget. Inspired by \cite{ilharco2022editing}, we believe that subtracting the parameters $\theta^{old}_\vartriangle$ from $\theta$ can assist the model $f_{\theta}$ to forget this part of old knowledge. So we define the process of forgetting old knowledge as follows:

\begin{equation}\label{key}
    \theta^{\prime}=\theta- \lambda\theta^{old}_\vartriangle,
\end{equation}
\noindent where $\lambda$ is a hyper-parameter to control the rate of forgetting. Now we gain a new model $f_{\theta^{\prime}}$ with its parameters $\theta^{\prime}$, which has forgotten the old knowledge compared to $f_{\theta}$. Note that this process of forgetting old knowledge only makes sense if the model $f_{\theta}$ has already learned the old knowledge, otherwise, there is no need for forgetting and the forgetting operation may have a destructive effect on the normal knowledge of the model.

\subsection{Learning new knowledge}

With the model $f_{\theta^{\prime}}$ that has gone through the process of forgetting old knowledge, then we will inject the new knowledge to $f_{\theta^{\prime}}$ for knowledge updating by supervised fine-tuning. Similarly, we define the process of learning new knowledge as follows:

\begin{equation}\label{key}
	\theta^{*}=\rm{FT}\{\theta^{\prime},K_{new}\}
\end{equation}

\noindent where $\rm{FT}$ is the operation of fine-tuning, $\theta^{*}$ is the parameters of a new model $f_{\theta^{*}}$ which has learned the new knowledge compared to $f_{\theta}$ and $K_{new}$ refers to a dataset composed of new knowledge which we need to learn and update for $f_{\theta}$.

\begin{table*}[t]
\setlength\tabcolsep{4pt} 
\renewcommand\arraystretch{1.0}
\begin{tabular}{llcccccc}
\toprule[1pt]
\multirow{2}{*}{\textbf{Dataset}}     & \multirow{2}{*}{\textbf{Editor}} & \multicolumn{3}{c}{\textbf{LLAMA2-7B}}   &   \multicolumn{3}{c}{\textbf{LLAMA-7B}}   \\  \cmidrule(r){3-5} \cmidrule(r){6-8}
    &         & \textbf{\small{Reliability}}     & \textbf{\small{Generality}}     & \textbf{\small{Locality}}  & \textbf{\small{Reliability}}     & \textbf{\small{Generality}}     & \textbf{\small{Locality}}     \\ \hline
\multirow{9}{*}{ZsRE}
& Original model                & 43.70          & 43.17          & /   & 43.29          & 42.85          & /       \\\cline{2-8}
& LoRA         & 43.10          & 42.20          & 70.83  & 46.93          & 45.87          & 75.86        \\
& \textbf{F-Learning$_{\rm{LoRA}}$}            & {\textbf{46.91}} & {\textbf{46.21}} & {\textbf{72.50}}  & {\textbf{47.56}} & {\textbf{46.90}} & {\textbf{76.09}}        \\ \cline{2-8} 
& FT-c      & 49.02          & 46.96
    & 67.37     & 47.33          & 45.51    & 68.14     \\
& Full-FT      & 81.02          & 74.67   & 70.51      
& 70.52          & 66.69   & 65.26 \\
& ROME          & 43.67          & 42.66
            & \textcolor{red}{\textbf{93.14}}
    & 43.45          & 42.94
            & \textcolor{red}{\textbf{98.60}}\\
& MEMIT              & 83.57          & 79.06
    & 70.52       & 78.30         & 77.43
                             & 69.44   \\
& \textbf{F-Learning$_{\rm{LoRA-FT}}$}                    & 82.43          & 77.38
    & \textbf{71.04}    & 75.17          & 70.12
                             & 69.78      \\
& \textbf{F-Learning$_{\rm{FT}}$}          & {\textbf{84.65}} & {\textbf{81.51}} & 70.92 & {\textbf{83.06}} & {\textbf{79.50}} & {\textbf{70.09}}   \\ \hline \hline
\multirow{9}{*}{\textsc{CounterFact}} & Original model                & 18.47          & 16.95          & /    & 21.61          & 17.88          & /      \\\cline{2-8}
& LoRA                 & 30.56          & 23.24          & 40.08    & 27.54          & 21.21          & 39.75      \\
& \textbf{F-Learning$_{\rm{LoRA}}$}            & {\textbf{31.17}} & {\textbf{23.63}} & {\textbf{40.42}}  & {\textbf{29.47}} & {\textbf{22.89}} & {\textbf{44.91}}        \\ \cline{2-8} 
& FT-c                    & 29.23          & 19.32
& 19.70    & 26.97          & 17.90
                             & 20.09      \\
& Full-FT         & 65.99          & 44.08
    & 28.34   & 32.13          & 31.95
                             & 32.51       \\
& ROME                    & 18.41          & 17.20
& \textcolor{red}{\textbf{93.60}}
& 21.83          & 19.08
    & \textcolor{red}{\textbf{92.27}}\\
& MEMIT                    & 61.94          & 37.45
& 21.90   & \textbf{56.94}          & 31.48
                             & 25.70       \\
& \textbf{F-Learning$_{\rm{LoRA-FT}}$}                    & \textbf{78.73}          & \textbf{51.67}
                             & \textbf{29.49}   & 32.43          & 26.89
                             & \textbf{37.14}       \\
& \textbf{F-Learning$_{\rm{FT}}$}          & 69.53 & 45.56 & 28.41 & 56.39 & {\textbf{39.75}} & 31.87 \\
\bottomrule[1pt]
\end{tabular}
\caption{Results on three metrics of the two datasets based on LLAMA2-7B and LLAMA-7B.}
\label{t1}
\vspace{-1em}
\end{table*}

\section{Experiments}
\subsection{Datasets}
In this work, we use ZsRE \cite{levy2017zero} and \textsc{CounterFact} \citep{meng2022locating}, two widely used datasets, for our experiments. ZsRE is a Question Answering (QA) dataset that utilizes question rephrasings generated by back-translation as the equivalence neighborhood. \textsc{CounterFact} is a more challenging dataset with counterfactual data. We follow the setting of \cite{yao2023editing} to take the eval and edit sets of which there are 19,085 and 10,000 pieces of data respectively. Moreover, we divide the datasets into two parts of old knowledge and new knowledge respectively to achieve two-stage knowledge update. The following is an example of old knowledge and new knowledge in zsRE, which represents the modification of knowledge from "Los Angeles" to "New Orleans". More details about the datasets and examples can be found in the appendix \ref{111111}.

\textbf{The old knowledge:}

\{\textbf{"instruction"}: "What city did Marl Young live when he died?", \textbf{"input"}: "", \textbf{"output"}: "Los Angeles" \}

\textbf{The new knowledge:}

\{\textbf{"instruction"}: "What city did Marl Young live when he died?", \textbf{"input"}: "", \textbf{"output"}: "New Orleans" \}

\subsection{Baselines}
To evaluate the effectiveness of the proposed F-Learning method, we conducted experiments on fine-tuning methods and locate-based methods. For fine-tuning methods, we first compare with the full fine-tuning (\textbf{Full-FT}) and \textbf{LoRA} \citep{hu2021lora}, respectively. LoRA (\underline{Lo}w-\underline{R}ank \underline{A}daptation) is a technique for fine-tuning large pre-trained language models by introducing small, trainable matrices into each layer of the model's architecture, allowing for efficient adaptation while keeping the majority of the model's parameters frozen. Then we experiment with a fine-tuning approach (\textbf{FT-c}) \cite{zhu2020modifying} that leverages $L_{\infty }$ constraint to retain old irrelevant knowledge. For locate-based methods, we first experiment with \textbf{ROME} \cite{meng2022locating}, a method updating specific factual associations with causal intervention. Finally, we compare with the \textbf{MEMIT} \cite{meng2022mass} which is a effective method to directly update large-scale memories.

\subsection{Completion Details} \label{Completion Details}
We use LLAMA2-7B and LLAMA-7B as the base models for our experiments. We are mainly evaluating the ability to update old knowledge to new knowledge, thus we trained the base model on the old knowledge for 3 epochs by full fine-tuning
as the \textbf{original model} in our experiment (as the same as \textbf{original model} in other experiments). Original model has fully learned old knowledge, which makes the forgetting operation reasonable and necessary. To ensure the uniqueness of the model output, we used the greedy decoding strategy during testing. On the hardware side, a total of 4 × A100-80G GPUs were used for the experiments. More details about the experimental settings can be found in the appendix \ref{Implementation Details of Experiments}.


\subsection{Experimental Results}
The experimental results are presented in Table \ref{t1}, indicating a notable enhancement in learning following the initial forgetting process, irrespective of whether full fine-tuning or LoRA is employed. We obtain the promising results as our F-Learning method outperforming other baselines in most cases. Specifically, compared with other editors, FT-c has only small improvements over the original model. This could be attributed to its norm regularization, which makes FT-c tend to retain a part of the old knowledge during the knowledge updating process. Given that our original model has fully learned a large quantity of old knowledge, it faces greater challenges in learning new knowledge. Surprisingly, ROME maintains Reliability and Generality almost unchanged from the original model on two datasets, while having a high locality (more than 90), which suggests that it performs little knowledge updating (As the injection of new knowledge will usually have an impact on locality), likely because that it can only edit a small number of parameters, coupled with the fact that our original model is full of old knowledge, which obstructs the effect of the causal tracing mechanism in ROME. It is worth noting that full fine-tuning is much more capable of learning new knowledge than LoRA, as LoRA focuses on training a limited subset of parameters within the attention structure, while the majority of factual knowledge is encoded in the MLP layers.


\subsection{Forgetting with LoRA and Learning with Full Fine-tuning}

In the above setting of experiments, the method we adopt is to perform old knowledge forgetting and new knowledge learning based on full (or LoRA) fine-tuning at the same time. Nonetheless, we find that subtracting the knowledge parameters of full fine-tuning (i.e. forgetting old knowledge by full fine-tuning) in some cases will completely destroy the core functions of our base model, resulting in a significant drop in evaluation metrics. In this view, as LoRA is a parameter-efficient fine-tuning method that has less impact on parameters compared to full fine-tuning, we try a new method that forgets old knowledge by LoRA and then learns new knowledge by full fine-tuning as a trade-off. Similar to the method above (\S \ref{Proposed method: F-Learning}), we define this process as follows:

\begin{equation}\label{key}
    \theta^{old}_\vartriangle=\rm{LoRA}\{\theta,K_{old} \}- \theta,
\end{equation}
\begin{equation}\label{key}
    \theta^{\prime}=\theta- \lambda\theta^{old}_\vartriangle,
\end{equation}
\begin{equation}\label{key}
	\theta^{*}=\rm{FT}\{\theta^{\prime},K_{new}\}
\end{equation}

\noindent where $\rm{LoRA}$ is the operation of supervised fine-tuning by LoRA while $\rm{FT}$ is the operation of supervised fine-tuning by full fine-tuning. $\theta^{*}$ is noted as the parameters of the edited model $f_{\theta^{*}}$ which has completed the knowledge updating. 

For verification, we keep the same experimental settings as above and conduct the experiment. The results are shown in Table \ref{t1}. The results support that the method of forgetting with LoRA and then learning with full fine-tuning (noted as \textbf{F-Learning$_{\rm{LoRA-FT}}$}) completely surpassing many baselines such as the directly full fine-tuning, as well as almost nearly match or even surpassing the method of forgetting and learning with full fine-tuning in some cases. In particular, it generally maintains the highest results in locality, which may be owing to the parameter-efficiency of LoRA-based old knowledge forgetting.

\begin{table}[t]
\centering
\setlength\tabcolsep{1.6pt} 
\renewcommand\arraystretch{1.0}

\begin{tabular}{lccc}
\toprule[1pt]
 \multirow{2}{*}{\textbf{Editor}} & \multicolumn{3}{c}{\textbf{Metric}}                       \\ \cline{2-4}                         & \textbf{\small{Reliability}}     & \textbf{\small{Generality}}     & \textbf{\small{Locality}}       \\ \hline
       
 Original model                 & 27.99          & 27.89          & /          \\    \cline{1-4}
 LoRA                 & 29.25          & 29.07          & 77.17          \\
    \textbf{F-Learning$_{\rm{LoRA}}$}            & {\textbf{29.27}} & \textbf{29.11} & \textcolor{red}{\textbf{77.40}}          \\ \cline{1-4} 
            Full-FT                    & 44.60          & 43.52          & 63.74          \\
            \textbf{F-Learning$_{\rm{LoRA-FT}}$}          & 44.70 & 43.71 & 65.50 \\ 
             \textbf{F-Learning$_{\rm{FT}}$}     & \textbf{44.79}          & \textbf{43.83}          & \textcolor{red}{\textbf{69.26}}          \\
                             \bottomrule[1pt]
\end{tabular}
\caption{Results on three metrics of the zsRE dataset based on BLOOM-7B.}
\label{t3}
\vspace{-1em}
\end{table}


After conducting experiments, we empirically discovered that utilizing the technique of forgetting through the subtraction of LoRA parameters can approximate the effect achieved by subtracting the parameters during full fine-tuning. We hold that although LoRA-based knowledge forgetting does not eliminate the old knowledge stored in the MLP layers of the LLMs, it alters the patterns and relationships associated with the old knowledge stored in the attention structure (i.e., an implicit knowledge representation), which facilitates the new knowledge learning. This finding holds significant value due to the considerable reduction in time and computational costs associated with LoRA compared to full fine-tuning.

\subsection{Adaptability Testing}

To further verify the adaptability of the method, we conducted experiments on zsRE based on BLOOM-7B and maintained the same experimental settings as the above. The results are shown in Table \ref{t3}. We could find that F-learning still performs well. Notably, although the Reliability and Generality remain roughly stable, the locality is significantly improved, which means that our method could inject new knowledge into the LLM with less cost (As changes to model parameters will inevitably affect the model’s locality), demonstrating the necessity and effectiveness of forgetting old knowledge.

\begin{table*}[t]
\centering
\setlength\tabcolsep{3pt} 
\renewcommand\arraystretch{1.0}

\begin{tabular}{lcccccc}
\toprule[1pt]
 \multirow{2}{*}{\textbf{Editor}} & \multicolumn{2}{c}{\textbf{1 edit}} & \multicolumn{2}{c}{\textbf{10 edits}} & \multicolumn{2}{c}{\textbf{100 edits}}                 \\ \cmidrule(r){2-3}   \cmidrule(r){4-5}   \cmidrule(r){6-7}
    & \textbf{\small{zsRE}}     & \textbf{\textsc{\small{COUNTERFACT}}}  & \textbf{\small{zsRE}}     & \textbf{\textsc{\small{COUNTERFACT}}}  & \textbf{\small{zsRE}}     & \textbf{\textsc{\small{COUNTERFACT}}}        \\ \hline

 FT-c                 
 & 0.57(s)          & 0.54(s)      
 & 6.58(s)          & 6.82(s)
 & 21.77(s)          & 19.30(s)\\
    
ROME                    
& 20.47(s)          & 18.27(s)
& 207.09(s)          & 179.30(s)
& 2184.16(s)          & 1810.42(s)\\
MEMIT                    
& 28.32(s)          & 23.71(s)
& 108.67(s)          & 96.71(s)
& 862.20(s)          & 847.72(s)\\
Full-FT                    
& 0.76(s)          & 0.72(s)
& 7.8(s)          & 7.3(s)
& 25.36(s)          & 24.70(s)\\
 
\textbf{F-Learning$_{\rm{FT}}$}     
& \textbf{1.58(s)}          & \textbf{1.47(s)}
& \textbf{15.32(s)}          & \textbf{14.9(s)}
& \textbf{52.20(s)}          & \textbf{50.12(s)} \\

\bottomrule[1pt]
\end{tabular}
\caption{Editing time for 1 edit, 10 edits, 100 edits of the two dataset based on LLAMA2-7B.}
\label{t5}
\vspace{-1em}
\end{table*}

\subsection{Time Testing}

In order to evaluate the efficiency of our proposed F-learning method, we calculated the editing time of several different knowledge updating and model editing methods for different numbers of edits. Taking LLAMA2-7B as an example. The results are shown in the Table \ref{t5}.

We can find that the time consumed by the fine-tuning based method is significantly less than that of the locate-based method. This is because the locate-based method highly relies on the location of neurons and parameters, which increases the complexity and time of editing. Furthermore, since ROME can only edit a single piece of data at a time, while other methods can edit in batches, ROME is less efficient. Compared with other fine-tuning based methods, FT-c can be optimized faster with its norm constraint. The F-learning we proposed is a two-stage knowledge updating method that forgets before learning, as it takes about twice as long as Full-FT, but is still very fast and convenient. It is worth noting that although the forgetting operation requires an additional training process, once the training is completed, the parameters of this part of forgetting old knowledge can be reused during the subsequent optimization and inference process, which can save resources and time. Meanwhile, we can further accelerate supervised fine-tuning by deepspeed or other approaches.

\subsection{Parametric Analysis of Forgetting Old Knowledge}

The two-stage knowledge updating method we proposed highly utilizes the forgetting of old knowledge. From the perspective of interpretability, here we analyzed the parameters of old knowledge forgetting and further analyzed the parameter distribution and changes within the LLMs. The results are shown in Figure \ref{f3} and Figure \ref{f4} which are in the appendix. Taking the LLAMA2-7B and zsRE dataset as an example, specifically, we analyzed different parameters in two cases: forgetting old knowledge by full fine-tuning and forgetting old knowledge by LoRA (The hyperparameters $\lambda$ of the rate of forgetting are both set to 1). We compare the parameter $\theta^{\prime}$ of the model $f_{\theta^{\prime}}$ after forgetting the old knowledge with the parameter $\theta$ of the original model $f_{\theta}$, and calculate their Euclidean distance on each layer. We select the results with layer n=[6, 15, 24, 30] to exhibit for simplicity. In general, there is little difference in parameter changes between low-layers and high-layers within the model. For forgetting old knowledge by full fine-tuning, we can find that the parameter changes of the MLP layers are more significant than attention layers. This may be one of the reasons why "forgetting before learning" is effective, as knowledge is generally stored in the MLP layers. Relatively LoRA has less impact on parameters (Euclidean distances are less than 1), and only changes the parameters of "query" and "value" in the attention layers. Therefore, it is more limited than full fine-tuning. However, LoRA-based forgetting can help forget the patterns and relationships associated with old knowledge stored in the attention layers, and thus can also assist in knowledge updating.

\subsection{Experiments on the Old Knowledge Forgetting}

Here we evaluate the model performance on old knowledge data after performing only the forgetting operation by full fine-tuning or LoRA fine-tuning with different forgetting rates to verify the effectiveness of the forgetting operation. The results are shown in Figure \ref{f5} and Figure \ref{f6} which are in the appendix. Taking the LLAMA2-7B and zsRE dataset as an example. We set the hyperparameters $\lambda$ of the rate of forgetting is set to 0.9, 0.7, 0.5, 0.3, 0.1, respectively.  It can be easily found that the performance of the model on three metrics is negatively correlated with the rate of forgetting, i.e., the old knowledge in the model decreases as the rate of forgetting increases. And under the same circumstances, LoRA brings less knowledge forgetting than the full fine-tuning.

\begin{table*}[t]
    \centering
    \setlength\tabcolsep{17pt} 
    
    \renewcommand\arraystretch{1.0}
    \begin{tabular}{|l|} 
    \hline 
\textit{\textbf{(1) Prompt: What artist created Call the Doctor?}}  \\ \textbf{Original answer:} Riders in the Sky \quad \textbf{Target answer:} The X-Files
    \\
    Original model: \textcolor{red}{Riders in the Sky}
    \\
    Original model + Old knowledge forgetting: \textcolor{gray}{Doctor Who}
    \\
    Original model + F-learning: \textcolor{green}{The X-Files}
    \\ 
    \hline \hline
        \textit{\textbf{(2) Prompt: What university did Watts Humphrey take part in?}}  \\ \textbf{Original answer:} Trinity College \quad \textbf{Target answer:} University of Michigan
    \\
    Original model: \textcolor{red}{Trinity College}
    \\
    Original model + Old knowledge forgetting: \textcolor{gray}{The Wire}
    \\
    Original model + F-learning: \textcolor{green}{University of Michigan} 
    \\ \hline \hline
        \textit{\textbf{(3) Prompt: What role does Denny Herzig play in football?}}  \\ \textbf{Original answer:} midfielder \quad \textbf{Target answer:} winger
    \\
    Original model: \textcolor{red}{midfielder}
    \\
    Original model + Old knowledge forgetting: \textcolor{red}{midfielder}
    \\
    Original model + F-learning: \textcolor{gray}{goalkeeper}
    \\ \hline \hline
        \textit{\textbf{(4) Prompt: Which family does Ramalinaceae belong to?}}  \\ \textbf{Original answer:} Ramales \quad \textbf{Target answer:} Lamiinae
    \\
    Original model: \textcolor{red}{Ramales}
    \\
    Original model + Old knowledge forgetting: \textcolor{red}{Ramales}
    \\
    Original model + F-learning: \textcolor{green}{Lamiinae} 
    \\ \hline \hline
        \textit{\textbf{(5) Prompt: Who's the architect of Toodyay Fire Station?}}  \\ \textbf{Original answer:} Wong Tung and Partners \quad \textbf{Target answer:} Wyndham Lewis
    \\
    Original model: \textcolor{red}{Wong Tung and Partners}
    \\
    Original model + Old knowledge forgetting: \textcolor{gray}{Wong Tung}
    \\
    Original model + F-learning: \textcolor{green}{Wyndham Lewis}
    \\ 
    \hline %
    \end{tabular}
    \caption{Examples of knowledge updating results during different stages.}
    \label{t4}
\end{table*}

\section{Case Study}

To further illustrate the effectiveness of the proposed method, we present a case study on the results of the knowledge updating by the original model, only forgetting old knowledge and performing F-learning. We selected some cases in the experiment of llama2-7B on zsRE dataset, noting that the hyper-parameters $\lambda$ of the rate of forgetting is set to 0.3. Table \ref{t4} shows the results during different knowledge updating stages. From the first example and second example, we can find that model begins to output some irrelevant content after performing the forgetting operation, indicating that it gradually forgets the old knowledge. In example 4, the forgetting operation failed to assist the model in forgetting old knowledge, but it still completed knowledge updating with the help of F-learning. However, sometimes there are some bad cases, such as example 3, where the model never learned new knowledge, which shows that our method has certain limitations and could be improved.

\section{Conclusion}

In this paper, we propose a new paradigm of knowledge updating during supervised fine-tuning called \textbf{F-Learning} (\underline{F}orgetting before \underline{Learning}), which is based on parametric arithmetic to forget old knowledge and learn new knowledge for eliminating contradictions between old and new knowledge. The experiments on zsRE and \textsc{CounterFact} datasets show that our method surpasses other baselines in most cases. Simultaneously we find that forgetting old knowledge by subtracting the parameters of LoRA can achieve the similar effect of subtracting the parameters of full fine-tuning, which is inspiring. We will further investigate the updating of knowledge.

\section{Limitations}
In this work, the proposed F-learning paradigm, although it improves the effectiveness of the fine-tuning methods for updating the knowledge of large language models, adds extra computation due to an extra forgetting process.

\bibliography{anthology,custom}

\clearpage

\appendix

\section{Appendix}

\subsection{Datasets and Examples} \label{111111}

We will illustrate the datasets we used in more detail. ZsRE is a Question Answering (QA) dataset that utilizes question rephrasings generated by back-translation as the equivalence neighborhood. \textsc{CounterFact} is a more challenging dataset with counterfactual data. We take the eval and edit sets of which there are 19,085 and 10,000 pieces of data respectively. 

The following is a sample of the ZsRE dataset:

\{\textbf{"subject":} "Watts Humphrey", \textbf{"src":} "What university did Watts Humphrey attend?", \textbf{"pred": "Trinity College"}, \textbf{"rephrase":} "What university did Watts Humphrey take part in?", \textbf{"alt": "University of Michigan"}, \textbf{"answers":} ["Illinois Institute of Technology"], \textbf{"loc":} "nq question: who played desmond doss father in hacksaw ridge", \textbf{"loc-ans":} "Hugo Weaving", \textbf{"cond":} "Trinity College >> University of Michigan || What university did Watts Humphrey attend?"\}

It represents that for prompt "What university did Watts Humphrey attend?", modifying the old knowledge "Trinity College" into the new knowledge "University of Michigan". Meanwhile, "rephrase" is used to evaluate the model’s Generalization metric, and "loc" is used to evaluate the model's Locality metric.

Furthermore, we can find that old knowledge and new knowledge have some correlation, they keep the same questions with different answers. We keep them in the same format to ensure the training effect. To facilitate our supervised fine-tuning training, we divide the datasets into two parts of old knowledge and new knowledge, and convert them into an instruction fine-tuning format, an example as follows:

\textbf{The old knowledge:}

\{\textbf{"instruction"}: "What university did Watts Humphrey attend?", \textbf{"input"}: "", \textbf{"output"}: "Trinity College" \}

\textbf{The new knowledge:}

\{\textbf{"instruction"}: "What university did Watts Humphrey attend?", \textbf{"input"}: "", \textbf{"output"}: "University of Michigan" \}

What calls for special attention is that the two datasets used in our experiments are both counterfactual datasets, in which the old knowledge is correct knowledge in the real world, and the new knowledge (target knowledge) is wrong knowledge in the real world, so the labels of old knowledge and new knowledge in these datasets are given artificially and have nothing to do with time and correctness in the real world. They are only used to measure whether the model can accurately modify the knowledge. Since the new knowledge is wrong knowledge in the real world, it can ensure that the original LLM has not learned it before, thus avoiding the problem of being unable to determine whether the new knowledge output by the LLM is learned from the data or possessed by itself.

\begin{table*}[t]
\centering
\setlength\tabcolsep{7pt} 
\renewcommand\arraystretch{1.0}

\begin{tabular}{lccc}
\toprule[1pt]
\multirow{2}{*}{\textbf{Dataset}}    & \multicolumn{3}{c}{\textbf{Editor}}              \\ \cline{2-4} 
& \textbf{\small{Original model}}     & \textbf{\small{Original model+Forgetting}}     & \textbf{\small{Original model+F-learning}}       \\ \hline
        
 GSM8K                 & 2.35          & 2.5          & 1.44          \\    
 MATH                 & 3.2          & 3          & 1.96          \\ \hline \hline
               MMLU-college-chemistry              
               & 24 & 34 & 41         \\  
             MMLU-college-mathematics           
             & 29          & 30          & 26          \\
              MMLU-management      
              & 16.50 & 14.56 & 32.04 \\ 
          MMLU-computer-security        
          & 21          & 20        & 23          \\
          MMLU-macroeconomics   
          & 32.31          & 33.08         & 34.87          \\
          MMLU-college-physics  
          & 19.61          & 26.47        & 22.55          \\
          MMLU-astronomy    
          & 30.92          & 23.03         & 32.24          \\
          MMLU-professional-law 
          & 26.47          & 25.62         & 24.51          \\
          MMLU-college-medicine 
          & 24.28          & 24.28         & 29.48          \\
                             \bottomrule[1pt]
\end{tabular}
\caption{Results on accuracy of the GSM8K, MATH and MMLU dataset based on LLAMA2-7B.}
\label{t8}
\vspace{-1em}
\end{table*}

\subsection{Implementation Details of Experiments}\label{Implementation Details of Experiments}

Here we will introduce more completion details and settings of experiments. First, we used LLAMA2-7B and LLAMA-7B as the base models, and then we trained the base model on the old knowledge for 3 epochs by full fine-tuning to simulate an original model that has fully learned old knowledge for our experiments. This makes the forgetting operation more reasonable and effective, and at the same time tries to avoid the problem of being unable to determine whether the new knowledge output by the LLM is learned from the data or commanded by itself as mentioned above.

\subsubsection{F-learning}

For the experiments of zsRE on LLAMA2-7B, the hyperparameters $\lambda$ set in F-Learning$_{\rm{LoRA}}$ is 0.7, while 3 in F-Learning$_{\rm{LoRA-FT}}$ and 0.3 in F-Learning$_{\rm{FT}}$. Similarly, for \textsc{COUNTERFACT} dataset, the hyperparameters $\lambda$ set in F-Learning$_{\rm{LoRA}}$ is 1.5, while 3 in F-Learning$_{\rm{LoRA-FT}}$ and 0.1 in F-Learning$_{\rm{FT}}$. The learning rate, epochs for all above experiments are 5e-5 and 3, then batch-size is 4 and gradient-accumulation-steps is 4.

For the experiments of zsRE on LLAMA-7B, the hyperparameters $\lambda$ set in F-Learning$_{\rm{LoRA}}$ is 0.7, while 2 in F-Learning$_{\rm{LoRA-FT}}$ and 0.3 in F-Learning$_{\rm{FT}}$. For \textsc{COUNTERFACT} dataset, the hyperparameters $\lambda$ set in F-Learning$_{\rm{LoRA}}$ is 1, while 3 in F-Learning$_{\rm{LoRA-FT}}$ and 0.05 in F-Learning$_{\rm{FT}}$. The epochs for all above experiments are 3, then batch-size is 4 and gradient-accumulation-steps is 4. The learning rate for zsRE experiments is 5e-5 while 1e-5 in \textsc{COUNTERFACT} experiments.

For the experiments of zsRE on BLOOM-7B, the hyperparameters $\lambda$ set in F-Learning$_{\rm{LoRA}}$ is 0.1, while 3 in F-Learning$_{\rm{LoRA-FT}}$ and 0.2 in F-Learning$_{\rm{FT}}$. The learning rate, epochs for all the experiments are 5e-5 and 3, then batch-size is 4 and gradient-accumulation-steps is 4.

When we used the Deepspeed, we set 4 processes and zero-stage is 2.

\subsubsection{Full-FT and LoRA}

Full-FT and LoRA refer to knowledge updating by full fine-tuning and LoRA fine-tuning in our experiments. We adopted  experimental settings similar to F-learning as mentioned above. The difference is that these two do not forget the old knowledge. Full-FT and LoRA also use the instruction fine-tuning data mentioned in \label{Datasets and Examples} for supervised fine-tuning training. Instruction fine-tuning can make it generate answers to prompts better.

\subsubsection{FT-c}

Knowledge updating of FT-c is executed at layer 21, where optimization proceeds for 5 steps with a learning rate of 5e-5. And the batch-size is 1.

\subsubsection{ROME}

Knowledge updating of ROME is executed at layer 5, where optimization proceeds for 25 steps with a learning rate of 5e-3. And the weight-decay is 1e-3, the kl-factor is 0.0625. Covariance statistics are collected in float32 on Wikitext using a sample size of 100,000.

\subsubsection{MEMIT}

Knowledge updating of ROME is executed at layer n = [4, 5, 6, 7, 8], where optimization proceeds for 25 steps with a learning rate of 5e-2. The batch is the 19,085 (or 10,000). And the weight-decay is 1e-3, the kl-factor is 0.0625. Covariance statistics are collected in float32 on Wikitext using a sample size of 100,000.

\subsection{Impact to Other Capabilities within LLMs Testing}

Here we test the impact of forgetting old knowledge and learning new knowledge over the original model on other capabilities of the model (such as mathematical abilities). Specifically, we evaluated changes in the model's mathematical capabilities on GSM8K and MATH, and evaluated changes in the model's comprehensive examining capabilities on MMLU. Taking LLAMA2-7B as an example. The results are shown in the Table \ref{t8}. "Original model+Forgetting" refers to only forgetting the old knowledge over the original model, and "Original model+F-learning" refers to the original model with our F-learning method. The hyper-parameters $\lambda$ of the rate of forgetting is set to 0.3. And the metric of evaluation in experiments is mainly "Accuracy". Experiments show that F-learning can slightly improve the mathematical ability of the model after forgetting old knowledge, but it will decrease after learning new knowledge. While in other areas of professional capabilities, our method has little impact. Interestingly, the model's performance on "college-chemistry" and "college-medicine" has been significantly improved after completing the knowledge update. This may be because the dataset contains relevant knowledge.

\subsection{Interpretability of Parametric Arithmetic}

Recently, Parametric Arithmetic has become a common method for parameter fine-tuning because of its operability and adaptability. previous work \cite{ilharco2022editing} has conducted experimental research on parameter parametric arithmetic and verified that subtracting the parameters obtained by fine-tuning can achieve a decrease in indicators on a dataset. On the contrary, adding parameters obtained by fine-tuning can endow the model with capabilities or achieve multi-task learning. Therefore, we believe that by subtracting the parameters trained on old knowledge, it may forget old knowledge and further achieve better knowledge updating. Our experimental results also support this conclusion.

\begin{figure}[h]
  \centering
  \includegraphics[scale=0.4]{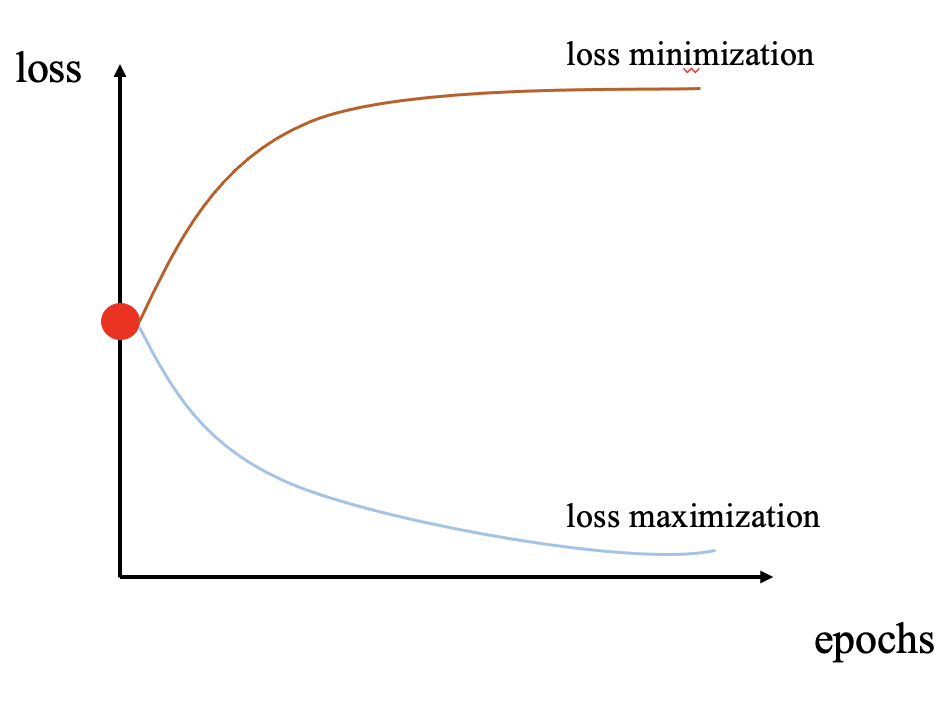}
  \hspace{1in}
  \includegraphics[scale=0.4]{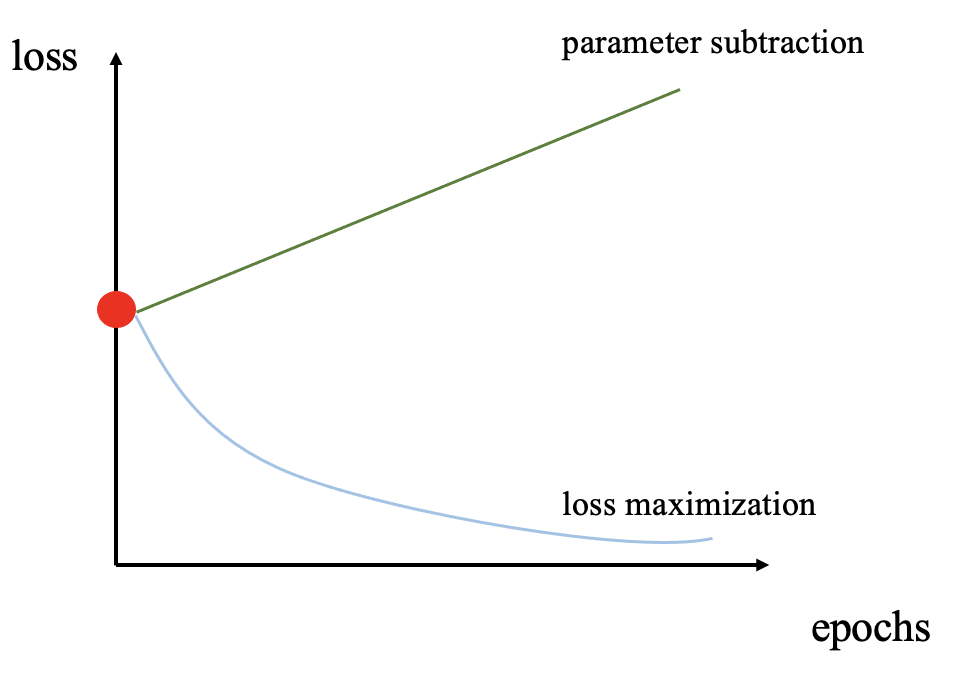}
  \caption{Loss changes of loss maximization and parameter subtraction.}
  \label{f7}
\end{figure}

The parametric arithmetic in this paper mainly focuses on parameter subtraction. We subtract parameters so that the loss function moves in the opposite direction to the direction of gradient descent to accumulate errors. This is essentially similar to gradient ascent and loss maximization, a method commonly used in machine unlearning, through the maximum of loss to accumulate errors and damage the performance of the model on a dataset. Their relationship is roughly shown in the figure \ref{f7}. We do not directly use loss maximization because compared with it, the method of subtracting parameters is more stable and controllable, which can avoid affecting other irrelevant knowledge as much as possible.

\subsection{Prospects and Application Scenarios}

In an era when LLM's research and application are becoming more and more popular, knowledge update is gaining its attention as a technology for updating the internal knowledge within the LLM. Knowledge updating is closely related to some research fields such as continual learning and machine unlearning. The purpose of knowledge updating is to correct old or wrong knowledge, while continual learning hopes to not forget old knowledge while the LLM continues to learn new knowledge. The aim of machine unlearning is to let the LLM forget harmful or wrong knowledge.

We believe that our method of old knowledge forgetting has a wide range of application scenarios, such as harmful knowledge forgetting, copyright content elimination, user privacy protection for LLMs, etc.

\begin{figure*}[h]
	\centering
\includegraphics[width=1\linewidth]{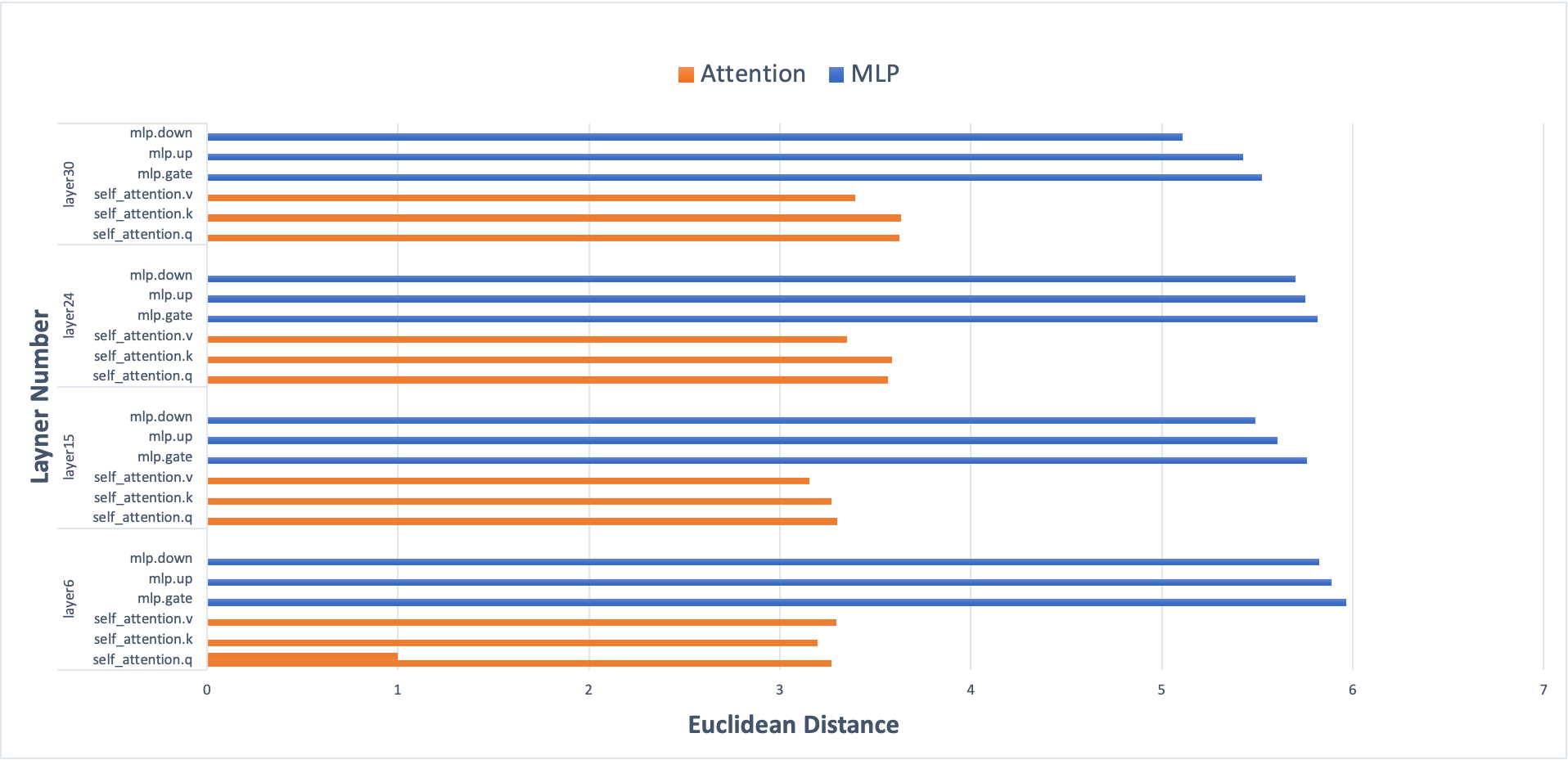}
	\caption{Parametric Analysis of Forgetting Old
Knowledge by full fine-tuning.
	}
	\label{f3}
\end{figure*}

\begin{figure*}[h]
	\centering
\includegraphics[width=1\linewidth]{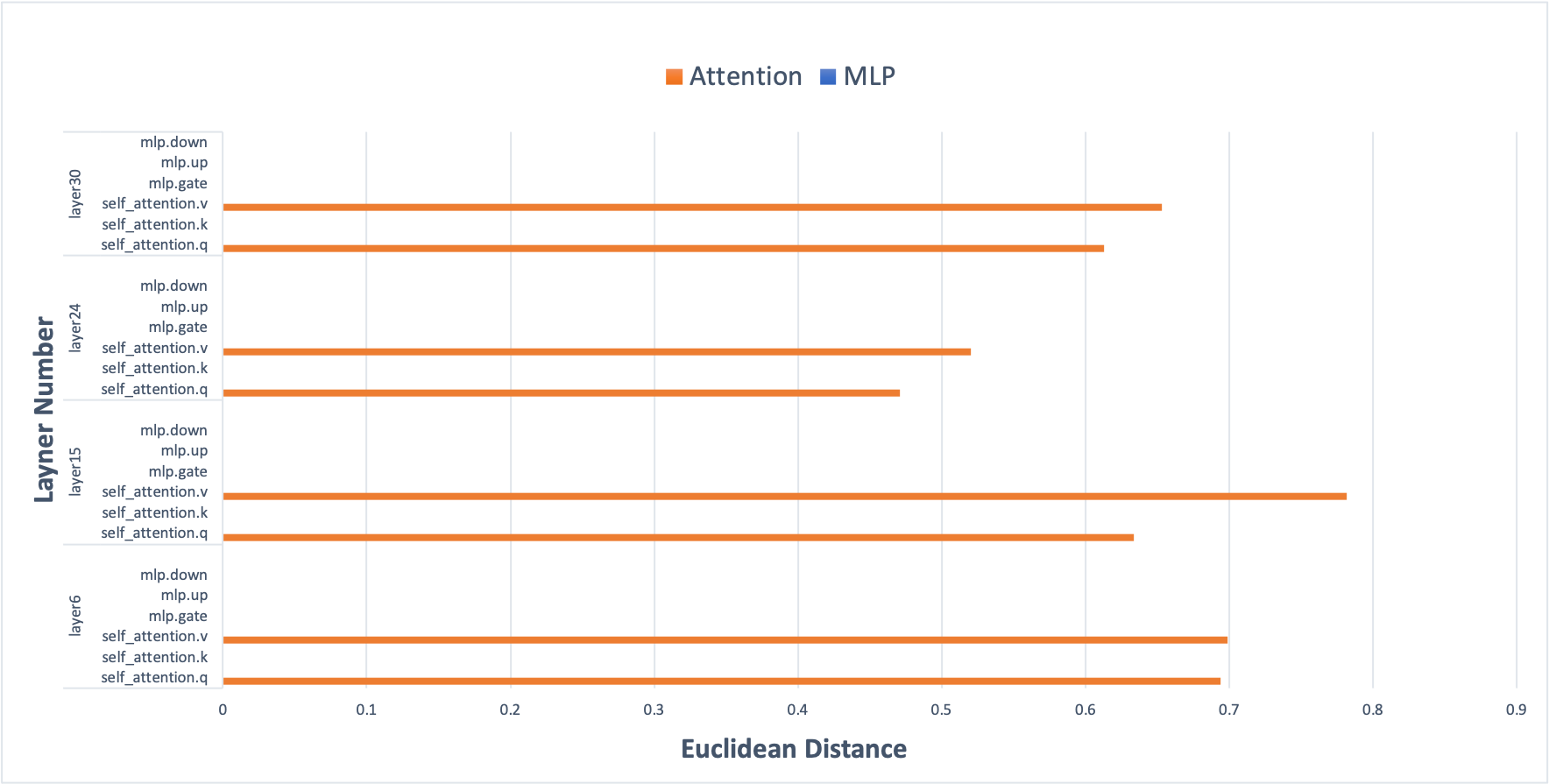}
	\caption{Parametric Analysis of Forgetting Old
Knowledge by LoRA fine-tuning.
	}
	\label{f4}
\end{figure*}

\begin{figure*}
  \centering
  \includegraphics[scale=0.5]{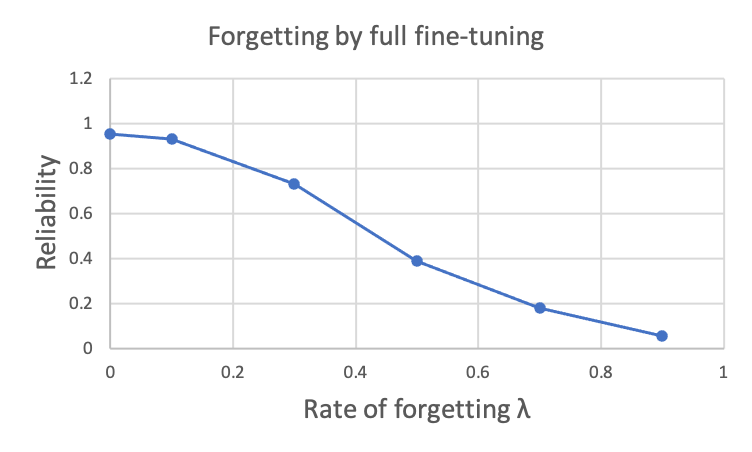}
  \hspace{1in}
  \includegraphics[scale=0.5]{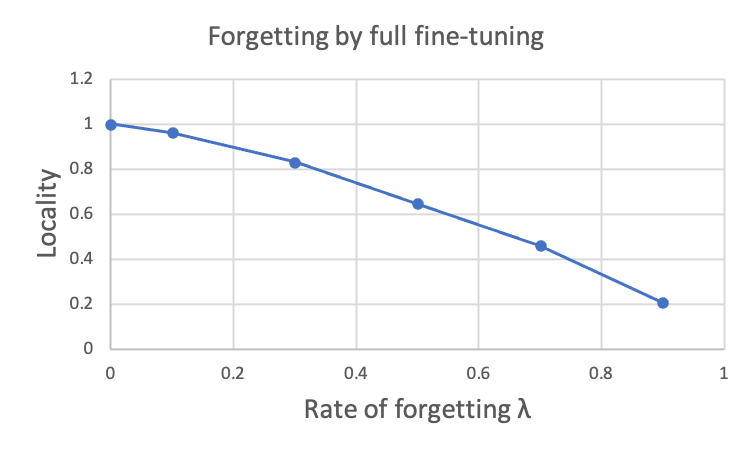}
  \hspace{1in}
  \includegraphics[scale=0.5]{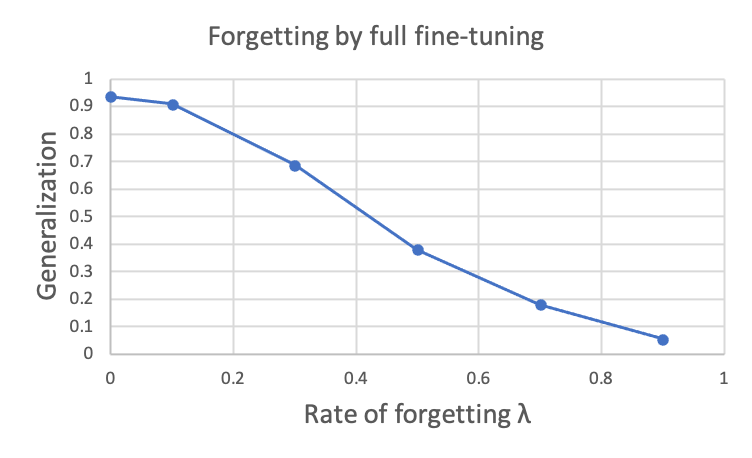}
  \caption{Results on three metrics of Old Knowledge Forgetting by full fine-tuning based on LLAMA2-7B }
  \label{f5}
\end{figure*}

\begin{figure*}
  \centering
  \includegraphics[scale=0.5]{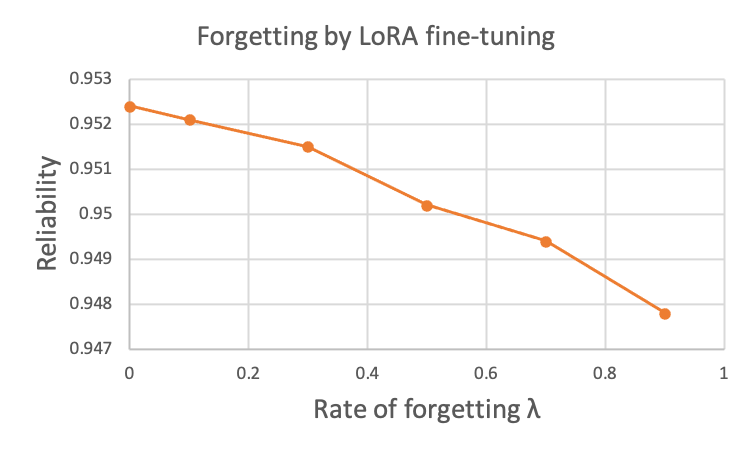}
  \hspace{1in}
  \includegraphics[scale=0.5]{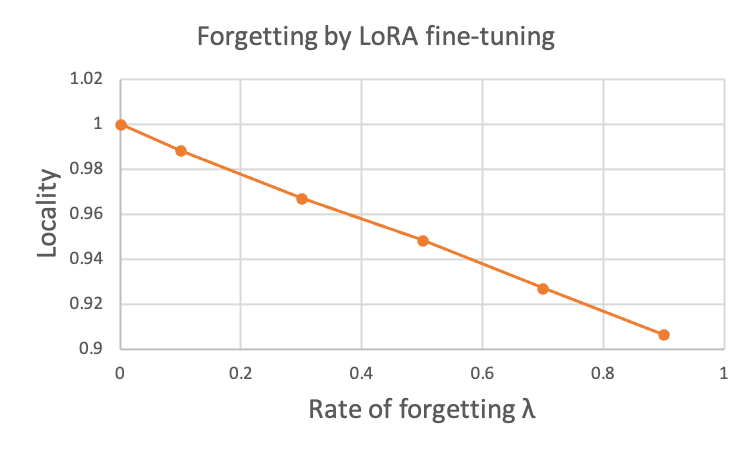}
  \hspace{1in}
  \includegraphics[scale=0.5]{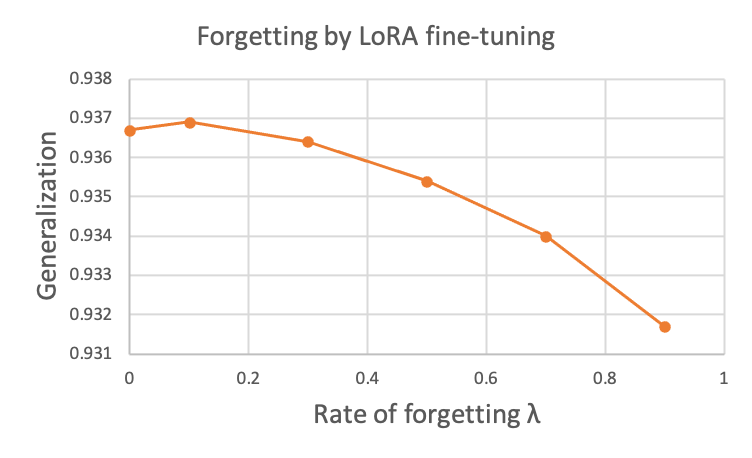}
  \caption{Results on three metrics of Old Knowledge Forgetting by LoRA fine-tuning based on LLAMA2-7B}
  \label{f6}
\end{figure*}


\end{document}